\documentclass[10pt,twocolumn,letterpaper]{article}

\usepackage{cvpr}
\usepackage{times}
\usepackage{epsfig}
\usepackage{graphicx}
\usepackage{amsmath}
\usepackage{amssymb}
\usepackage{caption}
\usepackage{subcaption}
\usepackage{bm}
\usepackage{multirow}
\usepackage{array}
\usepackage{url}

\usepackage{subcaption}
\usepackage{graphicx}

\usepackage{booktabs}
\usepackage{makecell}
\usepackage{pifont}

\usepackage{amsfonts}

\usepackage[pagebackref=true,breaklinks=true,letterpaper=true,colorlinks,bookmarks=false]{hyperref}

\cvprfinalcopy 

\def\cvprPaperID{xxxx} 

\ifcvprfinal\fi
\begin{document}

\title{SwinGar: Spectrum-Inspired Neural Dynamic Deformation for Free-Swinging Garments}

\author{Tianxing Li$^1$~~~~~~~~~~~Rui Shi$^1$~~~~~~~~~~~Qing Zhu$^1$~~~~~~~~~~~Takashi Kanai$^2$\\[0.3cm]
$^1$Beijing University of Technology, China\\[0.1cm]
$^2$The University of Tokyo, Japan\\[0.1cm]
{\tt\small litianxing@bjut.edu.cn}
}

\twocolumn[
{%
\renewcommand\twocolumn[1][]{#1}%
\maketitle

\begin{center}
    \includegraphics[width=\textwidth]{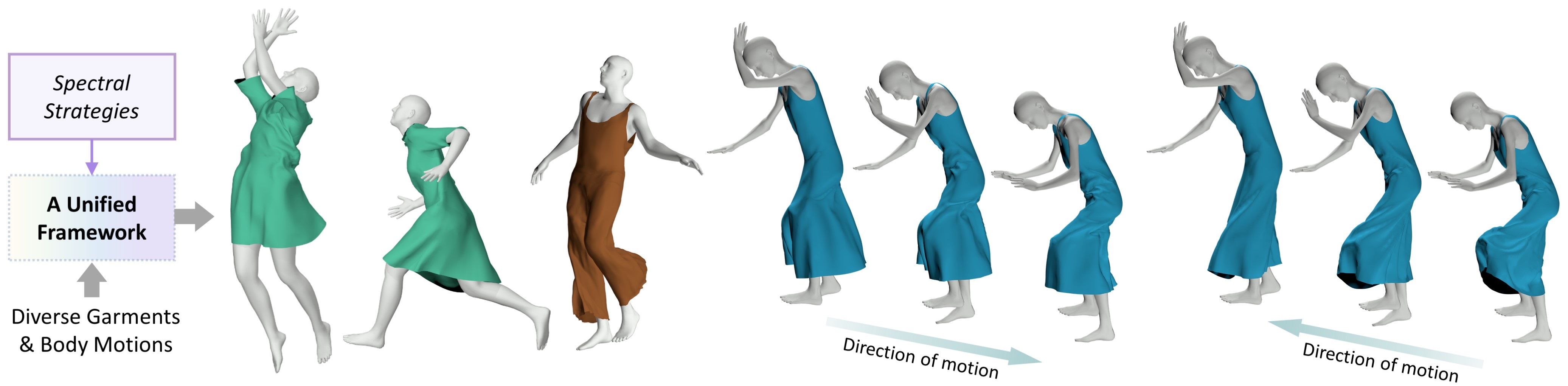}
    \captionof{figure}{We propose a learning-based method to automatically generate dynamic deformations across a variety of garments within a unified framework. By incorporating spectral strategies, our method achieves realistic generation of high-quality details while also exhibiting good generalization capabilities.}
    \label{fig:figTeaser}
\end{center}%
}
]


\begin{abstract}
  Our work presents a novel spectrum-inspired learning-based approach for generating clothing deformations with dynamic effects and personalized details. Existing methods in the field of clothing animation are limited to either static behavior or specific network models for individual garments, which hinders their applicability in real-world scenarios where diverse animated garments are required. Our proposed method overcomes these limitations by providing a unified framework that predicts dynamic behavior for different garments with arbitrary topology and looseness, resulting in versatile and realistic deformations. First, we observe that the problem of bias towards low frequency always hampers supervised learning and leads to overly smooth deformations. To address this issue, we introduce a frequency-control strategy from a spectral perspective that enhances the generation of high-frequency details of the deformation. In addition, to make the network highly generalizable and able to learn various clothing deformations effectively, we propose a spectral descriptor to achieve a generalized description of the global shape information. Building on the above strategies, we develop a dynamic clothing deformation estimator that integrates frequency-controllable attention mechanisms with long short-term memory. The estimator takes as input expressive features from garments and human bodies, allowing it to automatically output continuous deformations for diverse clothing types, independent of mesh topology or vertex count. Finally, we present a neural collision handling method to further enhance the realism of garments. Our experimental results demonstrate the effectiveness of our approach on a variety of free-swinging garments and its superiority over state-of-the-art methods. 
\end{abstract}

\section{Introduction}
Virtual humans, which are digital characters designed to resemble real humans, have been used in various industries. Realistic clothing is crucial for the appearance of virtual humans, making clothing animation an important topic in computer graphics.  Physics-based approaches  \cite{Adaptive18,minchen21}  apply basic physics laws to animate cloth, but they require extensive computation and are not practical for real-time applications. Alternately, learning-based methods \cite{GarNet19,GarNet22, FCGNN20} have been proposed to predict garment deformations close to simulation results, making them promising solutions for interactive cloth animation due to their efficiency.

So far, several learning-based methods \cite{TailorNet20,DeePSD21,PBNS21,GarFitNet23} have demonstrated the ability to generate plausible garment deformations under static poses. However, these models only consider isolated states in the current time, which results in a lack of dynamics in the animation, particularly for loose-fitting garments like dresses. More recently, researchers have also explored solutions for dynamic deformations using GRU temporal networks \cite{virBone22,Santesteban19}, inertia loss terms \cite{SNUG22,Hugo22}, or operating on image space  \cite{zhang21,zhang22}. However, these methods are restricted to specific garment objects and require individual training for each garment, which limits scalability.   

Undoubtedly, creating a unified learning-based model to approximate the behavior of animated garments is a difficult task due to the variety of clothing types, dynamic deformations, and nonlinear details involved. Neural networks with a carefully-designed architecture hold promise for this task, but they often display a bias towards low-frequency information, which has been criticized in other studies \cite{Hugo22,TailorNet20,GarNet22}. This bias becomes more pronounced when the model needs to learn diverse garment shapes. On the other hand, the common practice of directly using spatial position to represent clothing also poses a challenge for the task. This representation is highly sensitive and not generalizable, leading to overfitting and poor generalization performance of the model. Therefore, exploring solutions to these challenges is crucial.

We present a learning-based method for predicting the deformation of free-swinging garments. Our approach is centered on a general estimator based on graph attention mechanism and long short-term memory, capable of handling garments with arbitrary topologies.  The key to accomplishing such a complex task lies in the introduction of spectral analysis techniques, which facilitate effective control over the learning of low- and high-frequency garment deformations and the generation of discriminative global shape representations for diverse garments. With our designed estimator, high-quality predictions of dynamic deformations for unseen garments can be achieved without additional training. Besides the estimator for dynamic deformation, we also propose a novel collision handling method that can effectively remove collisions during inference while maintaining the reasonableness of garment details. This post-processing step enhances the overall visual appeal and realism of the clothing animation. Our main contributions include:  
\begin{itemize}
\item \textbf{A frequency control strategy for deformation learning.} We are the first to analyze the underlying nature of learning-based deformations lacking details from a spectral perspective. We introduce a frequency control strategy for graph attention layers to improve the deformation quality. 
\item \textbf{A spectral global clothing descriptor.} Considering the diverse range of customized garments in practice, we propose a spectral descriptor that provides robust and discriminative representation for various types of garments, allowing the model to effectively learn the personalized deformations relevant to garments.
\item \textbf{A general dynamic clothing estimator.} We present a novel estimator that leverages the combination of frequency-controllable graph attention mechanism with long short-term memory, enabling the generation of dynamic deformations for unseen garments without the need for repeated training.
\item \textbf{A neural collision-handling method.} We propose a collision detection and correction method using neural distance fields that can move clothing to a collision-free state while preserving its natural details.
\item \textbf{A neural collision-handling method.} We propose a collision detection and correction method using neural distance fields that can move clothing to a collision-free state while preserving its natural details.
\end{itemize}
Our proposed method has been validated through extensive experiments, demonstrating its advantages over state-of-the-art methods. The results highlight the potential of our approach to advance the field and facilitate practical applications, where the detailed insight into the ideas of learning-based deformation and spectral analysis can contribute to more future research.

\begin{figure*}[t]
\includegraphics[width=0.99\textwidth]{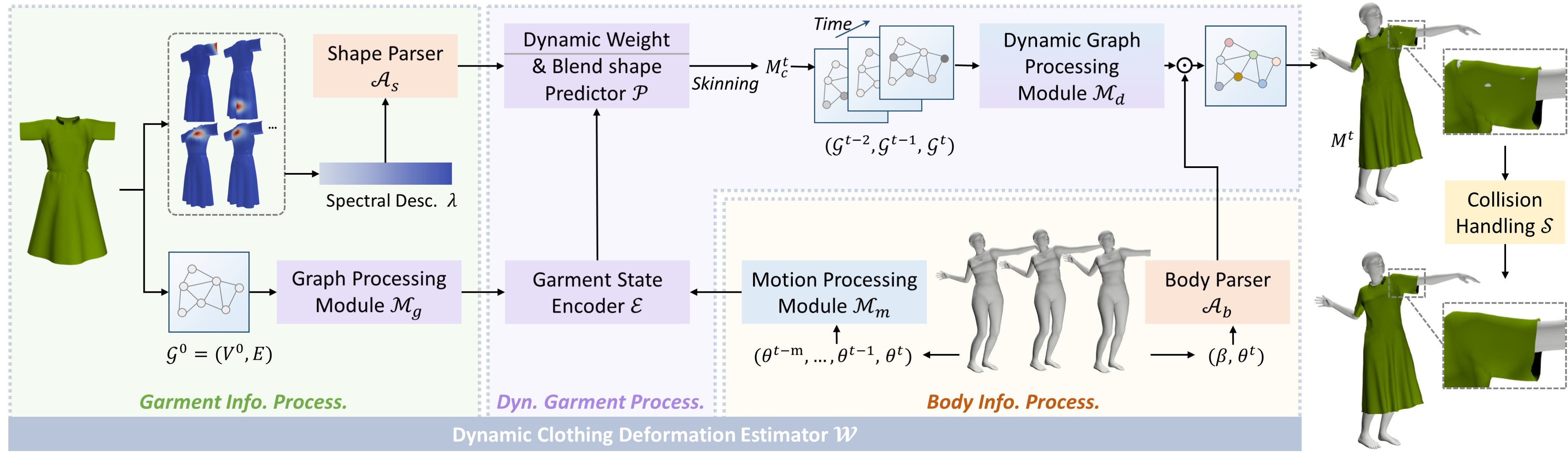}
\centering
\caption{Overview of our neural dynamic deformation method. Our framework consists of three units (\textit{i.e.}, garment information, body information, and dynamic garment processing) and a collision handling module. 
}  
\label{fig:figOverview}
\end{figure*}

\section{Related work}
\textit{Physics-based simulations} take into account the physical properties of clothing, allowing for a high degree of realism in deformation effects. The algorithm starts by constructing a model based on the physical properties and dynamics of the garment, and then numerically solves the model equations to obtain the deformed mesh state at each time step. Researchers have developed methods \cite{Jiang17,Adaptive18} to simulate hyper-realistic clothing, but the computational cost required for accurate results is enormous. Hence, studies have looked at ways to improve the efficiency and stability of garment simulation, including adding detailed wrinkles on low-frequency meshes \cite{Huamin21,zhen21}, utilizing parallel computation solutions with GPUs \cite{icloth18,pcloth20}, and trading some accuracy for performance in the approximation \cite{dry20}. On the other hand, physics-based simulations also face the challenge of setting simulation parameters. Typically, each change of cloth properties or resetting of the mesh structure requires manual fine-tuning of parameters, which requires a certain level of expertise and significant time cost. Despite the emergence of research \cite{Liang19} on the automatic acquisition of simulation parameters, there is still a need for computationally inexpensive and easier-to-set-up deformation methods if diverse garments are to be applied to digital scenes. 

\textit{Learning-based methods}  have gained popularity as an alternative to physics-based simulations \cite{Wang19,igor21,Ma20, ULNeF22,how23,PERGAMO22}. These methods use models for estimation and directly output the desired garment deformation. For complicated garments of game characters, NeuroSkinning \cite{Neuroskinning19} explored how to apply graph neural networks to 3D mesh deformation and proposed a skinning weight approximation method that can be applied to arbitrary topological structures of meshes. Subsequently, graph learning-based approaches \cite{rignet20,articulat21,DenseGATs20,li2021MultiResGNet} for automatic generation of skinning weights and blend shapes have been proposed. Although the above methods provide deformation approximation models with good generalization ability and efficiency, they are still limited to producing folds and wrinkles of garments with vivid visual expression. To this end, researchers have focused on garment deformation that generates rich details for dressing the parametric human body SMPL \cite{smpl15}. TailorNet \cite{TailorNet20} decomposes deformations under static pose into low- and high-frequency components, and then utilizes multiple multi-layer perceptrons (MLP) for deformation approximation, which can generate wrinkles for garment meshes with the fixed topology and number of vertices. To improve the generalization ability of deformation models, researchers use PointNet-based network \cite{GarNet19,GarNet22} and fully convolutional graph neural networks \cite{FCGNN20} to model deformations for various garments with different numbers of vertices. Similarly using the graph mesh-based network, N-Cloth \cite{nCloth22} can predict plausible garment deformation for arbitrary triangle meshes, while not being limited to SMPL bodies. Researchers have also made efforts to eliminate the dependency on huge volumes of ground truth data, by proposing a neural simulator trained by unsupervised learning \cite{PBNS21}. In addition, there are studies that follow an unsupervised scheme and introduce the inertia loss term, inspired by physics-based simulation, to achieve dynamic clothing effects \cite{SNUG22,Hugo22}. For loose-fitting garments, recent work \cite{virBone22} generates virtual bones for dresses and infers the dynamic deformation of garments from motion sequences. However, they are difficult to generate different dynamic garments using a single deformation approximation model. 

\section{Methodology}
\subsection{Frequency Control Strategy} \label{controlMethod}
Due to the wide range of garments with varying topologies and vertex counts, it is essential for our network to be capable of handling these diverse inputs.  To address this challenge, we adopt graph attention network (GAT) \cite{GAT18} and perform feature extraction using graph attention layers on arbitrary garment meshes. Let $V=[\boldsymbol v_1,...,\boldsymbol v_N] \in \mathbb{R}^{d \times N}$ denote the graph features, where $d$ is the dimension of features for each vertex, and $N$ is the number of vertices. The process of handling features in graph attention layer $f_{\rm Att}$ can be formulated as:
\begin{equation}
f_{\rm Att}(V)= V {\rm softmax}(g(V))^\top,
\end{equation}
where $g(V): \mathbb{R}^{d \times N} \rightarrow \mathbb{R}^{b \times N}$ denotes a linear transformation used to calculate attention scores and $b$ is the dimensionality of the feature after the transformation. The $\rm softmax$ function is employed as a normalization operation.

\begin{figure}[t!]
\includegraphics[width=1\linewidth]{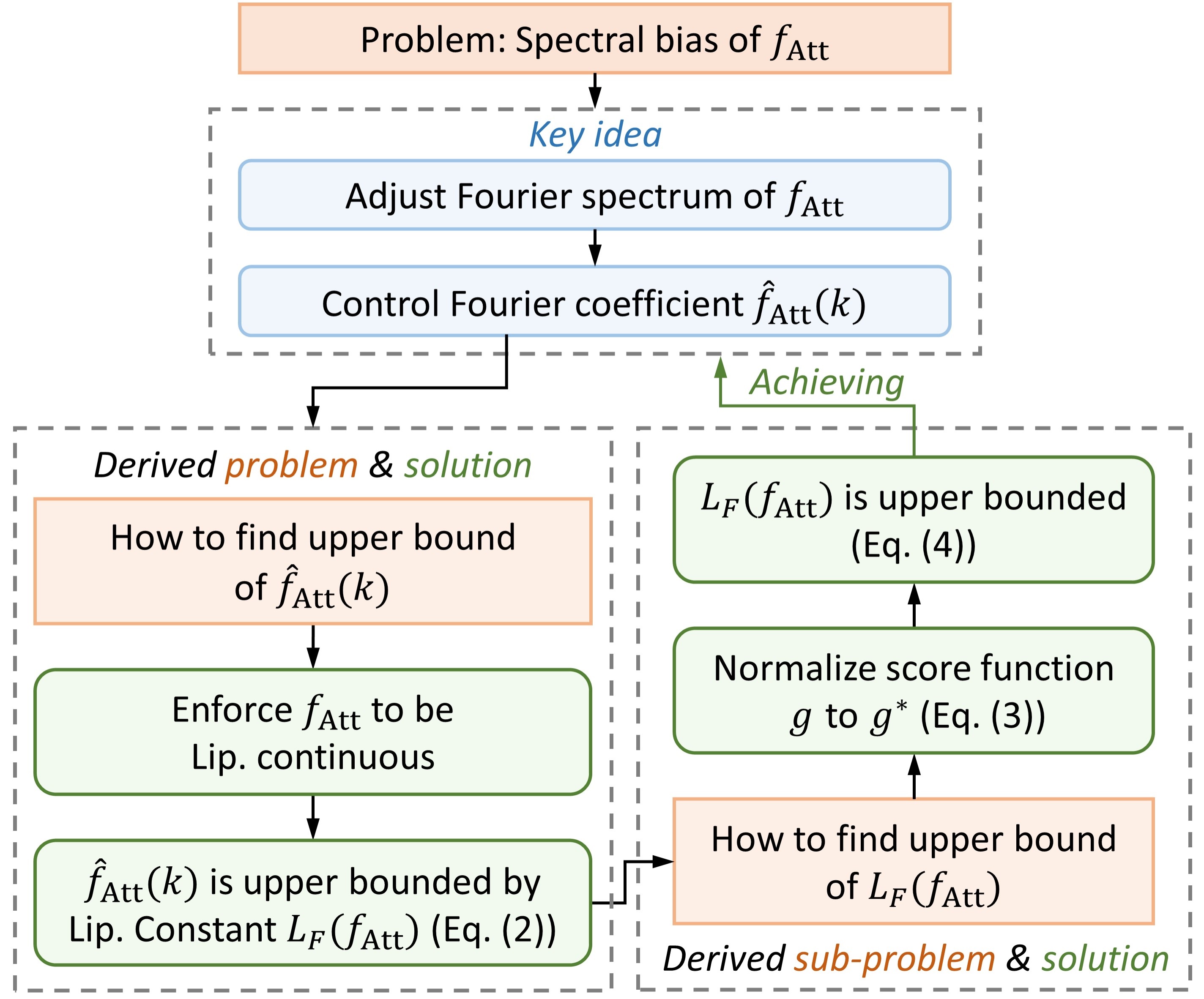}
\centering
\caption{Step-by-step procedure for addressing the spectral bias issue in graph attention layers from a Fourier spectrum perspective.}  
\label{fig:figLip}
\end{figure}

Graph attention-based networks have emerged as state-of-the-art methods in a wide range of 3D data processing applications. However, as previously argued, we have observed that these networks, along with other general graph networks, face challenges in learning deformations due to spectral biases. To address this issue, we aim to enhance the network's ability to learn high-frequency components by adjusting its spectrum (Fig. \ref{fig:figLip}). It has been demonstrated that finding an upper bound on the Fourier coefficients of the learning layer can help achieve control of the Fourier spectrum of the network \cite{bias22}. Therefore, the derived problem here becomes how to find the upper bound on the Fourier coefficients. To this end, according to the theory of harmonic analysis \cite{katznelson04}, we can enforce the graph attention layer $f_{\rm Att}$ to be Lipschitz continuous such that its Fourier coefficients are constrained. Specifically, if $f_{\rm Att}$ is Lipschitz continuous, there exists a constant $L$ satisfies $\|f_{\rm Att}(V_u)-f_{\rm Att}(V_w)\|_\mathrm{F} \leq L\|V_u - V_w\|_\mathrm{F}$ for any input of $V_u, V_w \in \mathbb{R}^{d \times N}$, where $\|\cdot\|_\mathrm{F}$ represents the Frobenius norm of the matrix. The minimum of such constant values is called the Lipschitz constant, denoted as $L_F(f_{\rm Att})$. 
Consequently, the $k$-th Fourier coefficient $\hat{f}_{\rm Att}(k)$ of the attention layer is bounded by:
\begin{equation}\label{eq2}
|\hat{f}_{\rm Att}(k)| \leq \pi \frac{L_F(f_{\rm Att})}{k}.
\end{equation}
Here, without any control of the attention layer, its Lipschitz constant $L_F(f_{\rm Att})$ is uncertain, which leads to no upper bound on the Fourier coefficient in Eq. \eqref{eq2}. At this point, a further derived problem arises: determining the appropriate upper bound for the Lipschitz constant $L_F(f_{\rm Att})$. Referring to \cite{lip2021}, we can normalize the attention score function $g(\cdot)$ such that $L_F(f_{\rm Att})$ is bounded. The normalized score function $g^*$ is defined as: 
\begin{equation}\label{eq3}
g^*(V)=\frac{\alpha g(V)}{{\rm max}\left\{\|g(V)\|_{(2, \infty)}, \|V^\top \|_{(\infty, 2)}L_{F,(2, \infty)}(g)\right\}},
\end{equation}
where $\alpha \geq 0$ is the scale controller of scores; $\|\cdot\|_{(2, \infty)}$ and  $\|\cdot\|_{(\infty, 2)}$ respectively denote $(2, \infty)$-norm and $(\infty, 2)$-norm for the matrix; $L_{F,(2, \infty)}(g)$ stands for the spectral norm of the parameters of $g$. The proof of Eq. \eqref{eq3} is detailed in \cite{lip2021}. With the designed normalization, $L_F(f_{\rm Att})$ is Lipschitz continuous, and Eq. \eqref{eq2} can be expressed as: 
\begin{equation}\label{eq4}
|\hat{f}_{\rm Att}(k)| \leq \pi \frac{L_F(f_{\rm Att})}{k} \leq \pi \frac{e^{\alpha} \sqrt{b/N}+ \alpha \sqrt{8}}{k}.
\end{equation}
As a result, the attention layer $f_{\rm Att}$ with $g^*$ is Lipschitz continuous, and its Fourier coefficient $\hat{f}_{\rm Att}(k)$ can be upper-bounded. This allows us to have control over the network's ability to learn high-frequency information through the parameter $\alpha$.  We also refer to the attention layer that has the score function $g^*$ described above as the frequency-controllable attention (FCA) layer $f_{\rm FCA}$.

\subsection{Clothing Spectral Descriptor} \label{discriptorMethod}
Existing dynamic garment deformation models \cite{virBone22, Hugo22} are mostly limited to the specific garment, while static garment deformation models that use graph neural networks \cite{nCloth22,DeePSD21} attempt to learn across multiple garment types but often rely only on vertex positions, which overlooks important global information and leads to inaccurate predictions.  Research has demonstrated that spectral shape analysis is effective in capturing mesh information \cite{HodgeNet21}. In this work, we investigate the use of spectral descriptors to reveal meaningful global information about garments with varying shapes.

Given a garment mesh with $N$ vertices, we construct an affinity matrix $A \in \mathbb{R}^{N \times N}$, where the $i, j$-th entry of $A$ is the affinity between the $i$-th and the $j$-th vertices. Specifically, we use the geodesic distance calculated by Dijkstra's algorithm to define the affinity between two vertices. The defined affinity matrix has the advantage of being invariant to transformations and bending, allowing for the effective learning of diverse garment deformations. Next, we perform spectral decomposition on the affinity matrix $A$. The eigenvectors of the affinity matrix form the normalized representation of the garment shape, while the eigenvalues specify how the shape varies along the axes. This allows us to consider the eigenvalues as the spectral descriptor.

\begin{figure}[t!]
  \centering
  \includegraphics[width= 0.99\linewidth]{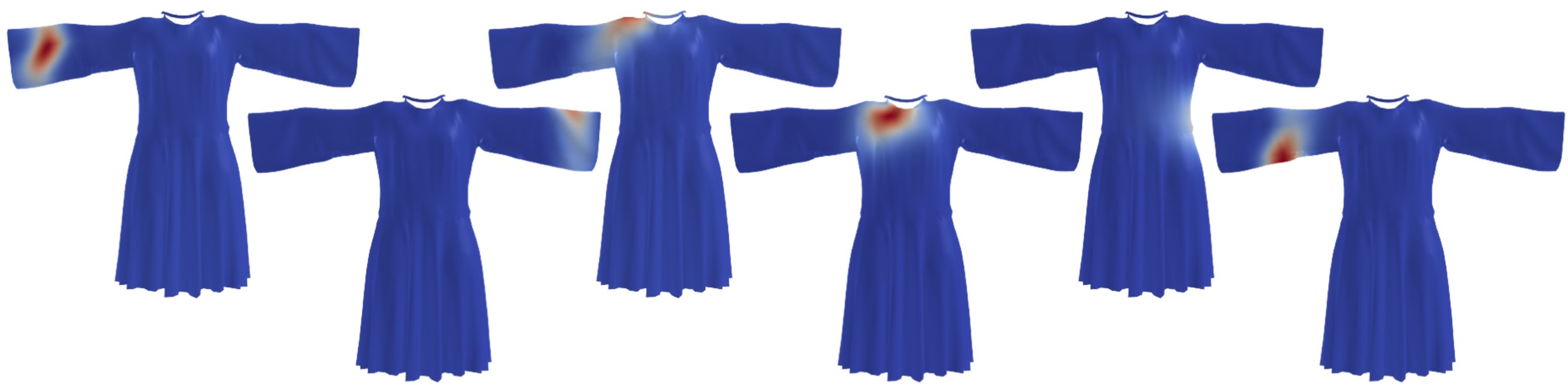}
  \caption{\label{fig:figSpecFeaVec} Color plots of the first six approximated eigenvectors of the long-sleeved dress mesh. 
}
\end{figure}

In practice, garments often have thousands or tens of thousands of vertices, which can make the process of constructing and decomposing affinity matrices computationally expensive. To expedite this process, we apply Nyström approximation method \cite{Nystrom04} to efficiently approximate the eigenvalues of the original affinity matrix $A$. In particular, we perform furthest point sampling, a technique where we start by randomly selecting a vertex and then iteratively sample from the remaining vertices that are farthest from the set of already-sampled vertices until $z$ vertices are selected ($z \ll N$). For these sampled vertices, we then calculate the affinity matrix $B \in \mathbb{R}^{z \times z}$. This smaller affinity matrix can be easily spectral-decomposed $B=U \Lambda U^\top$, allowing us to obtain an approximation of the eigenvectors $\hat Q$ of the original affinity matrix by using the Nyström method. Then, the corresponding affinity matrix can be approximated as $\hat A=\hat Q \Lambda \hat Q^\top$. Finally, we combine the eigenvalues into a vector $\boldsymbol \lambda = [\lambda_1, \lambda_2, \ldots, \lambda_z]$, which we refer to as our clothing spectral descriptor. Note that the length $z$ of the designed spectral descriptor is determined by the number of samples selected from the original garment shape and remains fixed. In contrast to the direct representation of clothing shapes by vertex coordinates, the proposed spectral descriptor offers a more compact and global representation of the garment shape. This enables the model to efficiently learn multiple types of garment deformations. To demonstrate a clearer visualization of the global information description, we show a color plot of the approximated eigenvectors $\hat Q$ (instead of the eigenvalues, which are difficult to represent visually) in Fig. \ref{fig:figSpecFeaVec}. 

\subsection{Dynamic Clothing Deformation Estimator} \label{estimatorMethod}
Given an arbitrary 3D garment mesh in the initial state $M^0$, a set of SMPL \cite{smpl15} bodies with shape parameters $\beta$ and continuous poses $(\theta^{t-m},...,\theta^t)$ from the previous state $t-m$ to the current state $t$, our goal is to predict the deformed garment mesh with dynamic effects based on the state of the body and the properties of the clothing itself. Formally, we define our dynamic deformation estimator $\mathcal{W}$ as:
\begin{equation}
M^t = \mathcal{W}(M^0, \beta, (\theta^{t-m},...,\theta^t)).
\end{equation}
Fig. \ref{fig:figOverview} illustrates the dynamic deformation approximation process using the estimator $\mathcal{W}$. Due to the multi-source nature of the deformation-related information, the estimator is divided into several units responsible for processing garment information, body information, and dynamic garment. More specifically, each unit contains the network structure designed to handle different features, including a garment processing module $\mathcal{M}_g$, a state encoder $\mathcal{E}$, a dynamic weight and a blend shape predictor $\mathcal{P}$ with FCA layers; a motion processing module $\mathcal{M}_{m}$ with LSTM layers; a dynamic graph processing module $\mathcal{M}_{d}$ with LSTM-FCA layers; a shape parser $\mathcal{A}_{s}$, a body parser $\mathcal{A}_{b}$, a collision handling model $\mathcal{S}$ with fully-connected layers. 

\textit{Garment information processing.} To extract adequate garment information, we first construct a mesh graph for initial garment $\mathcal{G}^0=(V^0, E)$, where $V^0$ represents the vertex features and $E$ denotes the mesh edges. For each vertex, we define its features as: $\boldsymbol v_i=[n_i, x_i, a_i]^\top$, which includes  vertex normal $n_i \in \mathbb{R}^3$, position $x_i \in \mathbb{R}^3$, and the garment-body fit attribute $a_i \in \mathbb{R}^1$ \cite{GarFitNet23}. These features capture important characteristics of the garment and describe the fit to the target body. We then pass this mesh graph through a graph processing module $\mathcal{M}_g$ to extract high-dimensional local graph features. On the other hand, we use spectral global description to analyze the garment as a whole, using the descriptor $\boldsymbol \lambda$. This descriptor is then fed into a shape parser $\mathcal{A}_{s}$ to extract the comprehensive shape information. 

\textit{Body information processing.} The garment deformation is also closely influenced by body state. Therefore, we adopt a motion processing module $\mathcal{M}_{m}$ composed of LSTM to sequentially processes the motion poses $(\theta^{t-m},...,\theta^t)$. In practice, we set $m$ to 64 to include the relative overall motion information. Additionally, we apply a body parser $\mathcal{A}_{b}$ to project the body shape $\beta$ and current pose $\theta^t$ to the latent space, resulting in the latent body representation.

\textit{Dynamic garment processing.} To encode the dynamic garment state, we perform an element-wise multiplication between the processed graph features from the graph processing module $\mathcal{M}_{g}$ and the processed motion features from the motion processing module $\mathcal{M}_{m}$, and feed the product into the garment sate encoder $\mathcal{E}$ to extract garment state features. Next, loose garments such as dresses require skinning weights that are strongly correlated with both the clothing shape and the motion. To achieve this, we integrate garment shape and state information by multiplying them together and then generate garment and motion-dependent skinning weights $W^t$ through a dynamic weight predictor $\mathcal{P}$.  This predictor also internally incorporates part of the branching structure, enabling the simultaneous generation of the blend shape $B^t$.  Subsequently, using the resulting skinning weights and blend shapes,  we apply linear blend skinning to generate the intermediate deformation $M_c^t={\rm LBS}((M^0+B^t),\theta, W^t)$. Note that this $M_c^t$ is order-dependent and incorporates underlying dynamics resulting from dynamic weights and blend shapes.

\begin{figure}[t]
  \centering
  \includegraphics[width= 0.99\linewidth]{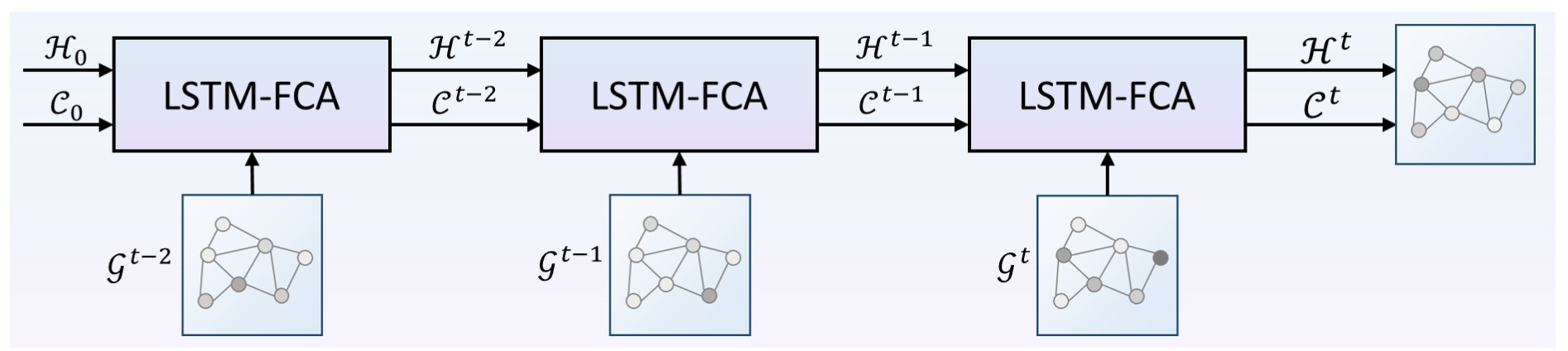}
  \caption{\label{fig:figLSTMGAT} Illustration of the inside of dynamic graph processing module. Sequences of graphs are transformed using the LSTM, while the state-to-state and input-to-state transformations follow the FCA mechanism. 
}
\end{figure}

To enhance the detailed folds of garments and ensure continuity, it is essential to process the time-series garment mesh data further. To do so, we construct a mesh graph $\mathcal{G}^{t}=(V^t, E)$ based on $M_c^t$, where each vertex feature includes the current position of the vertex and its distance from all body skeletal joints. We then use a dynamic graph processing module $\mathcal{M}_{d}$ to map a series of graphs to corrective detail adjustments. For efficiency, we use three consecutive graphs $(\mathcal{G}^{t-2},\mathcal{G}^{t-1},\mathcal{G}^{t})$. The design of $\mathcal{M}_d$ integrates the frequency-controlled attention mechanism into the original LSTM:  
\begin{align} \begin{aligned}
&i^t = \sigma \left(f_{\rm FCA}(\mathcal{G}^t)+f_{\rm FCA}(\mathcal{H}^{t-1})+w_i\mathcal{C}^{t-1}\right),\\
&f^t = \sigma\left(f_{\rm FCA}(\mathcal{G}^t)+f_{\rm FCA}(\mathcal{H}^{t-1})+w_f\mathcal{C}^{t-1}\right),\\
&o^t = \sigma\left(f_{\rm FCA}(\mathcal{G}^t)+f_{\rm FCA}(\mathcal{H}^{t-1})+w_o\mathcal{C}^{t-1}\right),\\
&\mathcal{C}^t = i_t\circ	\tanh\left((f_{\rm FCA}(\mathcal{G}^t)+f_{\rm FCA}(\mathcal{H}^{t-1})\right)+f^t \circ \mathcal{C}^{t-1},\\
&\mathcal{H}^t = o^t \circ \tanh \left(\mathcal{C}^t\right),
\end{aligned} \end{align}
where $i^t, f^t, o^t$ denote the input, forget, and output gate respectively; $\mathcal{C}^t, \mathcal{H}^t$ indicate the cell state and hidden state; $w_i, w_f, w_o$ are the weights of the cell state; $\sigma$ is the sigmoid function, and $\circ$ means the Hadamard product. The structure of $\mathcal{M}_d$ is shown in Fig. \ref{fig:figLSTMGAT}. The designed module considers both the spatial correlation between structural graph nodes and the temporal correlation between sequential graphs, enabling efficient learning of high-level representations from complex clothing graph data. Next, to make the dynamic details fully reflect the attributes of the garment and the body, the processed graph after the dynamic graph processing module $\mathcal{M}_{d}$ is multiplied with the latent body features from the body parser $\mathcal{A}_{b}$ to generate the final detail correction. This correction is at the vertex level and will be added to the intermediate garment $M_c^t$ to obtain the dynamic garment deformation $M^t$.

To train the dynamic clothing deformation estimator, besides the position loss, we also employ consistency and collision loss terms \cite{PBNS21} to help learn garment deformations close to the ground truth, while ensuring continuity and collision-free. 

\subsection{Collision Handling}\label{collisionMethod}
Despite the collision loss term provides some soft constraints on penetration between the garment-body pair during model optimization, it may not be effective in handling collisions for unseen data during inference. To address this issue, we propose a neural collision-handling method that can accurately detect and appropriately respond to garment collisions, correcting the garments to be collision-free while preserving realistic details.

Neural fields are capable of representing the physical properties of the object across space, and have been successfully applied in several 3D tasks \cite{ULNeF22,field22}. For the potential collision between our dynamic garments and bodies, we use the signed distance field (SDF) of the body to represent the penetration between the body and the garment:
\begin{equation}\label{eqSDF}
s(x_i)= {\rm sgn}(x_i, M_b) d(x_i, M_b),
\end{equation}
where $d(x_i, M_b)\in \mathbb{R}$ is the unsigned nearest distance from garment vertex $x_i$ to body mesh $M_b$, and $\rm sgn (\cdot)$ is a sign function indicating whether the $x_i$ is inside (positive) or outside (negative) of the body. 


The function of Eq. \eqref{eqSDF} is non-analytic, and the computation for each vertex can be time-consuming. Therefore, we propose employing a neural model $\mathcal{S}$ to approximate the SDF for fast collision detection. The neural network $\mathcal{S}$ comprises multiple fully connected layers. To facilitate effective parameter optimization, we design a loss function that guides the learning process: 
\begin{align}
\mathcal{L}_{\rm SDF}= &\frac{1}{N} \sum_{i=1} ^{N} \left( \|\mathcal{S}(x_i)-s(x_i)\| +  \|\nabla_{x} \mathcal{S}(x_i) - n_i\|  \right)  \nonumber \\
    &+ \mathcal{L}_{\rm E},\\
&\mathcal{L}_{\rm E} = \frac{1}{N} \sum_{i=1} ^{N}  (\|\nabla_{x} \mathcal{S}(x_i)\|-1)^2, 
\end{align}
where $\|\cdot\|$ is the $L_2$ norm. The loss function encourages the prediction $\mathcal{S}(x_i)$ to be similar to the ground truth value $s(x_i)$ and its gradients $\|\nabla_{x} \mathcal{S}(x_i)\|$ to be similar to the normal $n_i$. Also,  we use an Eikonal term \cite{Implicit20} $\mathcal{L}_{\mathrm E}$ that constrains the predicted gradient value $\|\nabla_{x} \mathcal{S}(x_i)\|$  to be close to 1. $\mu_e$ is the balancing weight and set as 0.15.

Once a garment vertex collision has been quickly detected by the model, we need to make reasonable adjustments to the position of the collision vertex. For the collision vertex $x_i$ with a positive value of $\mathcal{S}(x_i)$, the corrective displacement $\Delta(x_i) \in \mathbb{R}^3$ is defined as:
\begin{equation}
\Delta(x_i)= \frac{\nabla_{x_i}  \mathcal{S}(x_i)}{\|\nabla_{x_i} \mathcal{S}(x_i)\|}\left(|\mathcal{S}(x_i)|+ \delta_i\right),
\end{equation}
where $\nabla_{x_i} \mathcal{S}(x_i)$ is normalized as the direction of displacement, and the magnitude is calculated as the sum of two terms. The first term of the magnitude is the SDF value $|\mathcal{S}(x_i)|$ (i.e. the collision vertex $x_i$ is moved just to the body boundary) and the second term $\delta_i$ is a further detail correction based on the state of the non-collision neighboring vertices $\mathcal{N}_i$. Specifically, $\delta_i=\sum_{k \in \mathcal{N}_i}w_k |\mathcal{S}(x_k)|/{|\mathcal{N}_i|}$ indicates the weighted average of the absolute SDF value $|\mathcal{S}(x_k)|$ of the adjacent vertex $x_k \in \mathcal{N}_i$, where the weight $w_k$ is determined by the Gaussian function applied to the distance between $x_i$ and $x_k$, which is then normalized. Here, $\delta_i$ serves two purposes: first, it allows for partial avoidance of edge-to-edge collisions that may occur if only the vertices were moved to the surface of the body using the magnitude of $|\mathcal{S}(x_i)|$; second, it helps maintain local consistency near the corrected vertex, preventing excessive bulges. 

\section{Experiments}
We obtained loose garments, including long t-shirts, dresses of varying lengths and sleeve styles, and jumpsuits, from the public CLOTH3D dataset \cite{cloth3d}, and draped them over different SMPL bodies. To animate these characters wearing various garments, we collected motion sequences (\textit{e.g.}, dancing, running, throwing, strutting, etc.) from the CMU mocap dataset \cite{CMU} at a frame rate of 30, then utilized a physics-based simulator to create ground truth data. The training set consists of about 50 garments with 50,000 poses, while the test set includes 15 different garments with 7,500 poses, and there is no overlap between them. More implementation details for training and testing are presented in the supplemental materials.

\subsection{Results and Evaluation}
\begin{figure}[t!]
  \centering
  \includegraphics[width= 0.99\linewidth]{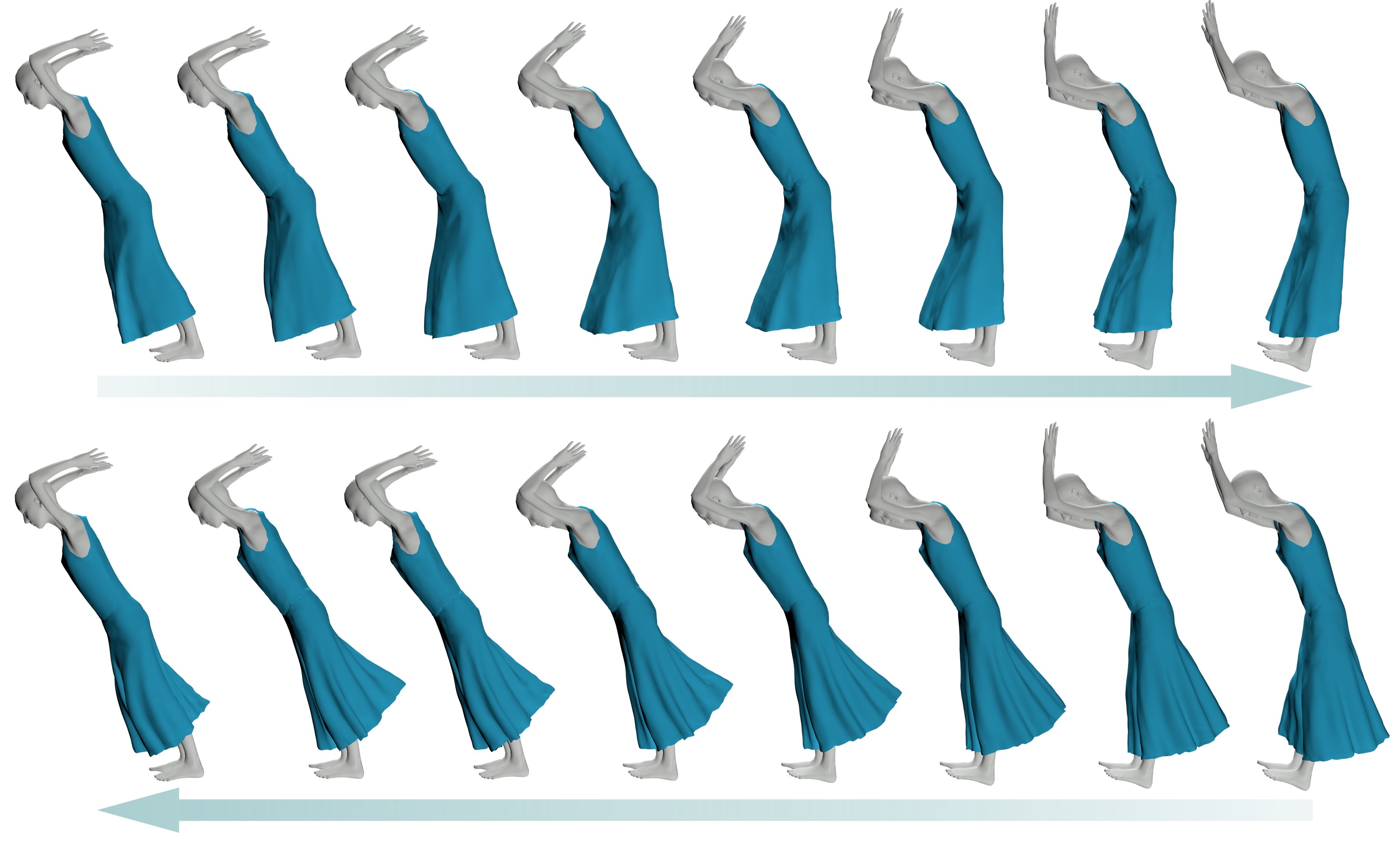}
  \caption{\label{fig:figDynamic} Example of a forward and backward motion sequence of dressing swinging. Our method can successfully generate dynamic effects.
}
\end{figure}
\textit{Dynamic effect.} In Fig. \ref{fig:figDynamic}, given the dressing and swinging motions from the test set, our approach is capable of predicting the unseen garment's dynamic effect. The trend of body movement can be clearly observed by examining the positions of the head and arms. In the top row, the body undergoes a swing from a forward-leaning state to a slightly curved state, resulting in the front hemline of the garment transitioning from a hanging state to a position close to the calf. In contrast, in the bottom row, the body motion is from back to front, causing the back side of the dress to gradually lift up with the movement. Despite the same pose, our dynamic deformation estimator successfully predicts the distinct garment dynamics based on the previous states. More intuitive dynamic results are available in the supplementary material.

\begin{figure}[t!]
  \centering
  \includegraphics[width= 0.99\linewidth]{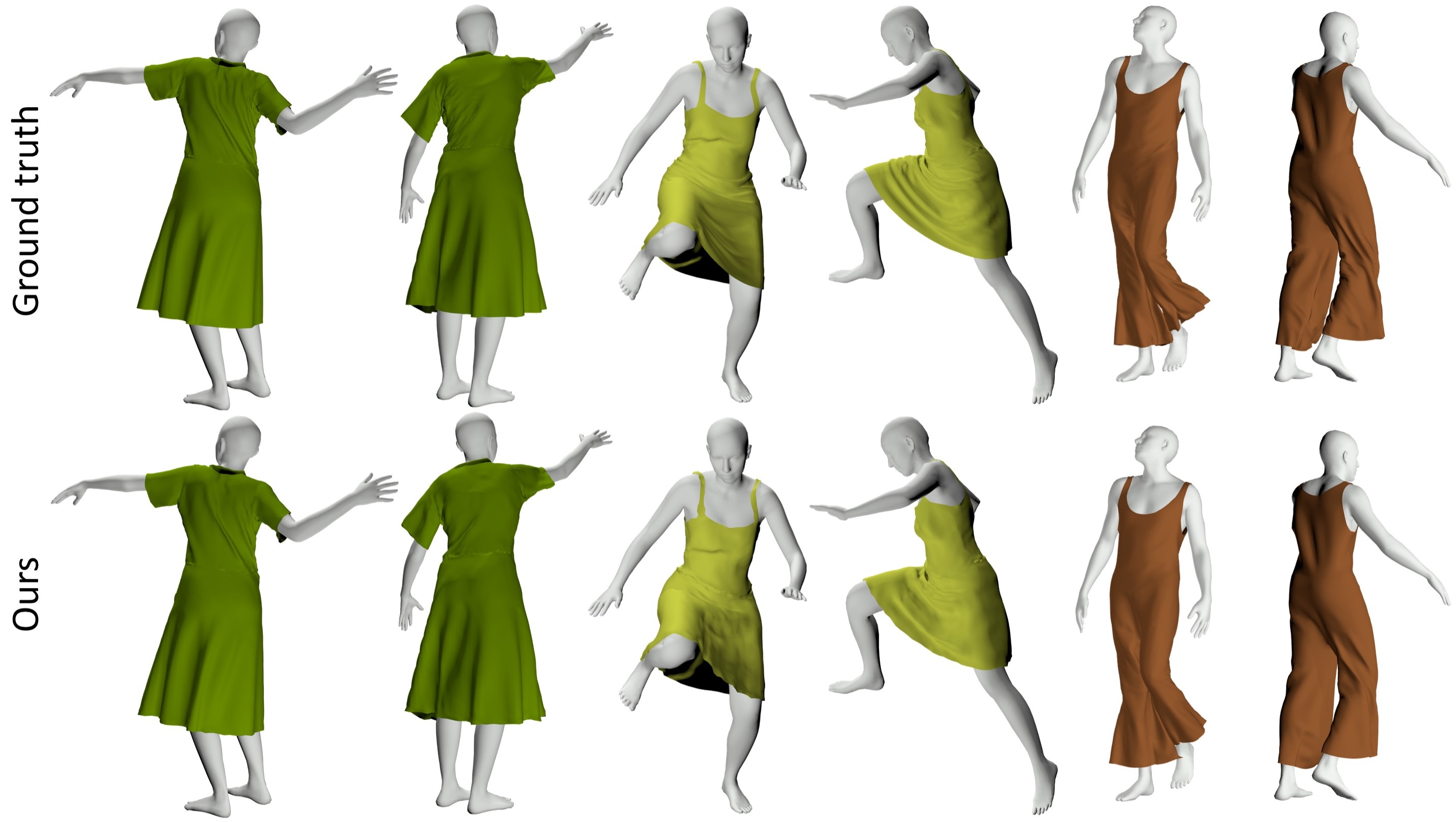}
  \caption{\label{fig:figReality} Qualitative results of our method for various garments and motions. 
}
\end{figure}
\textit{Realism.} In Fig. \ref{fig:figReality}, we present the deformation realism performance of our method on diverse garment types and motions. It should be noted that all of these test garments and motions are unseen simultaneously during training. While some subtle folds in the back, hemline, and trouser legs may not be fully generated, the overall deformation details are realistic and exhibit a high degree of fidelity to the ground truth. Moreover, our method is able to learn garment dynamics through a unified model. This eliminates the need for repeated training for different garments, making the framework highly efficient and compatible. 

\begin{figure}[t!]
  \centering
  \includegraphics[width= 0.78\linewidth]{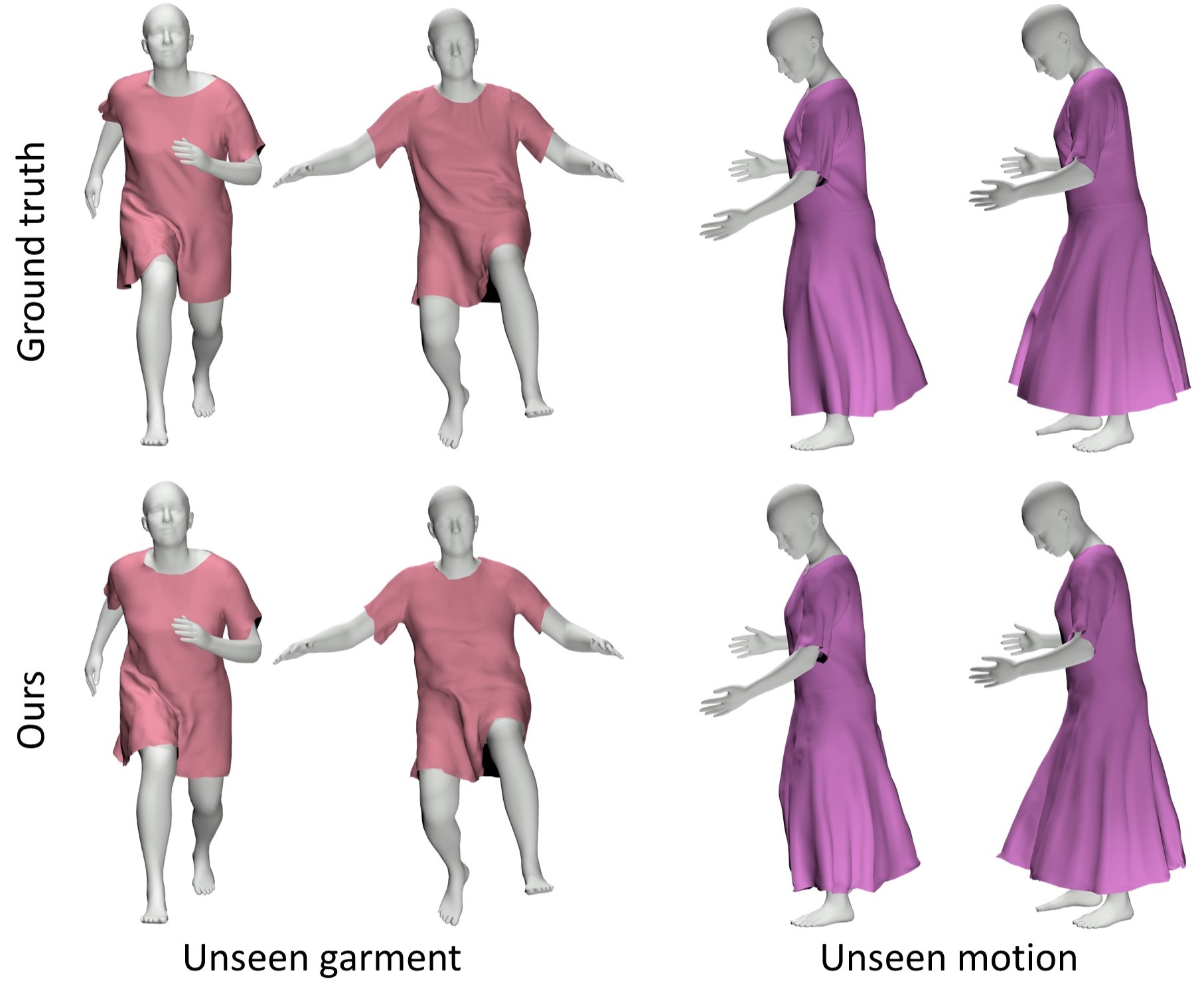}
  \caption{\label{fig:figGeneralization} Generalization to new garments and new motions respectively. 
}
\end{figure}

\textit{Generalization.} Fig. \ref{fig:figGeneralization} shows the quantitative results of generalization to a new garment and a new motion. In particular, we evaluate the proposed method on a new garment, \textit{i.e.}, the loose long t-shirt, which has different mesh connectivity and number of vertices compared to the training samples. Our method can produce plausible deformation results with similar fold trends and dynamic behavior without requiring repeated training, thanks to the spectral strategies and the designed deformation estimator. Moreover, when predicting with a new motion, our framework effectively captures spatio-temporal information through the careful design of the LSTM module for body movements and the LSTM-FCA for temporal graphs. This results in an approximate match between the predicted dress hemline and ground truth, showcasing the ability of our method to reasonably generate garment dynamics even with unseen motions.

\subsection{Ablation Study}
\begin{figure}[t!]
  \centering
  \includegraphics[width= 0.98\linewidth]{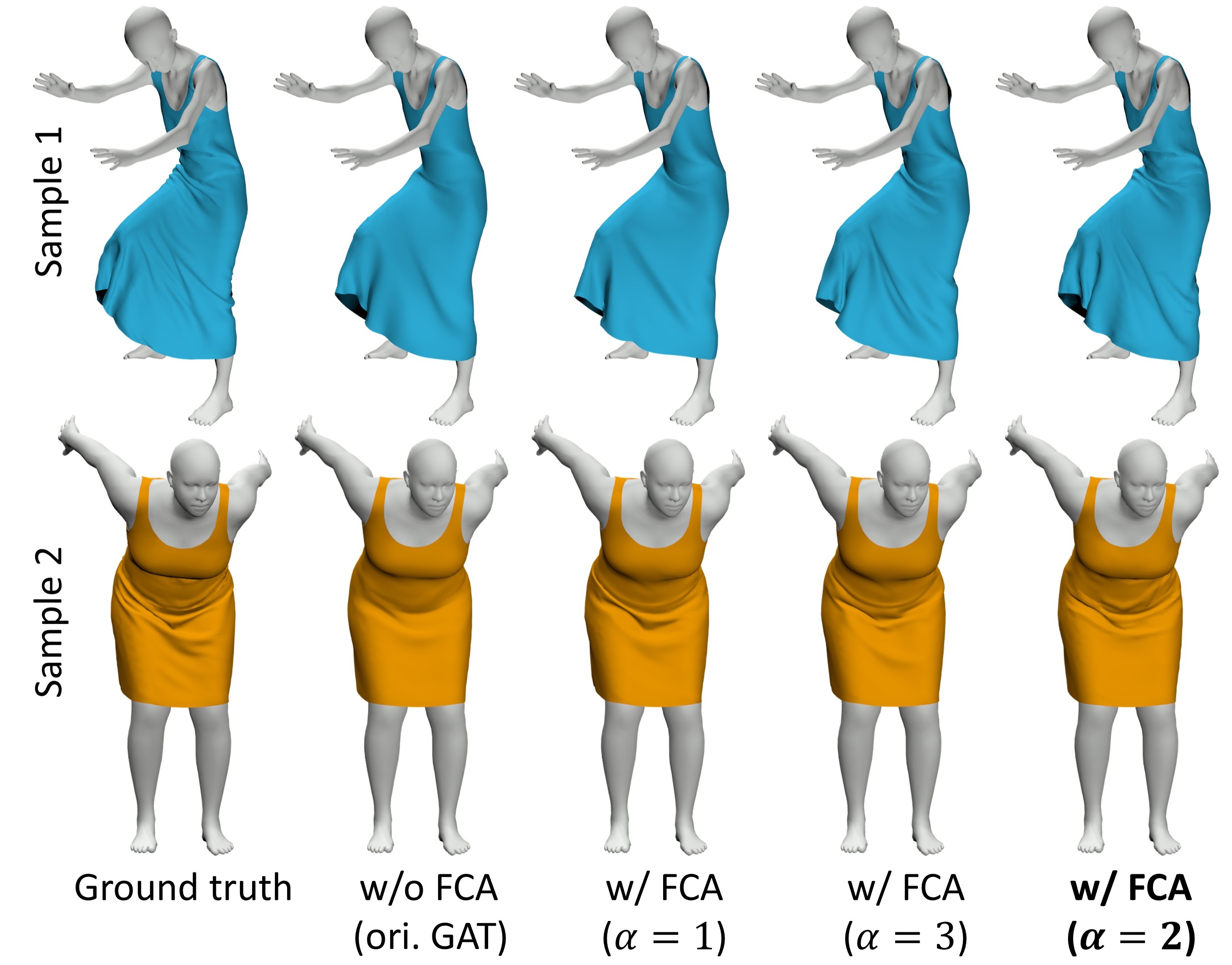}
  \caption{\label{fig:figAblaFCA} Qualitative results of ablation on frequency controllable attention layers.}
\end{figure}

\begin{figure}[t!]
     \centering
     \begin{subfigure}[c]{0.23\textwidth}
         \centering
         \includegraphics[width=\textwidth]{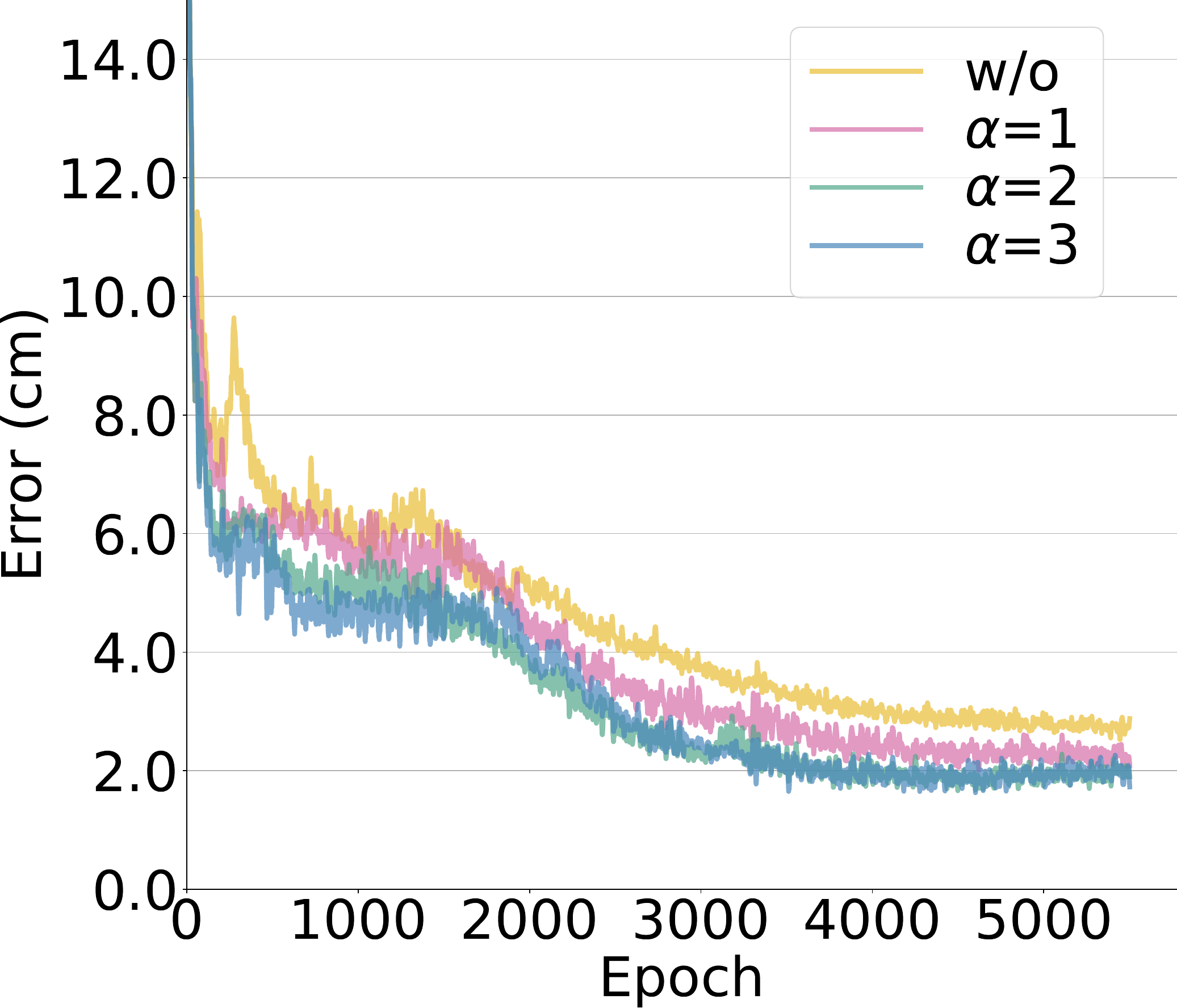}
         \caption{}
         \label{fig:figAblaFCALoss}
     \end{subfigure}
     \hfill
     \begin{subfigure}[c]{0.23\textwidth}
         \centering
         \includegraphics[width=\textwidth]{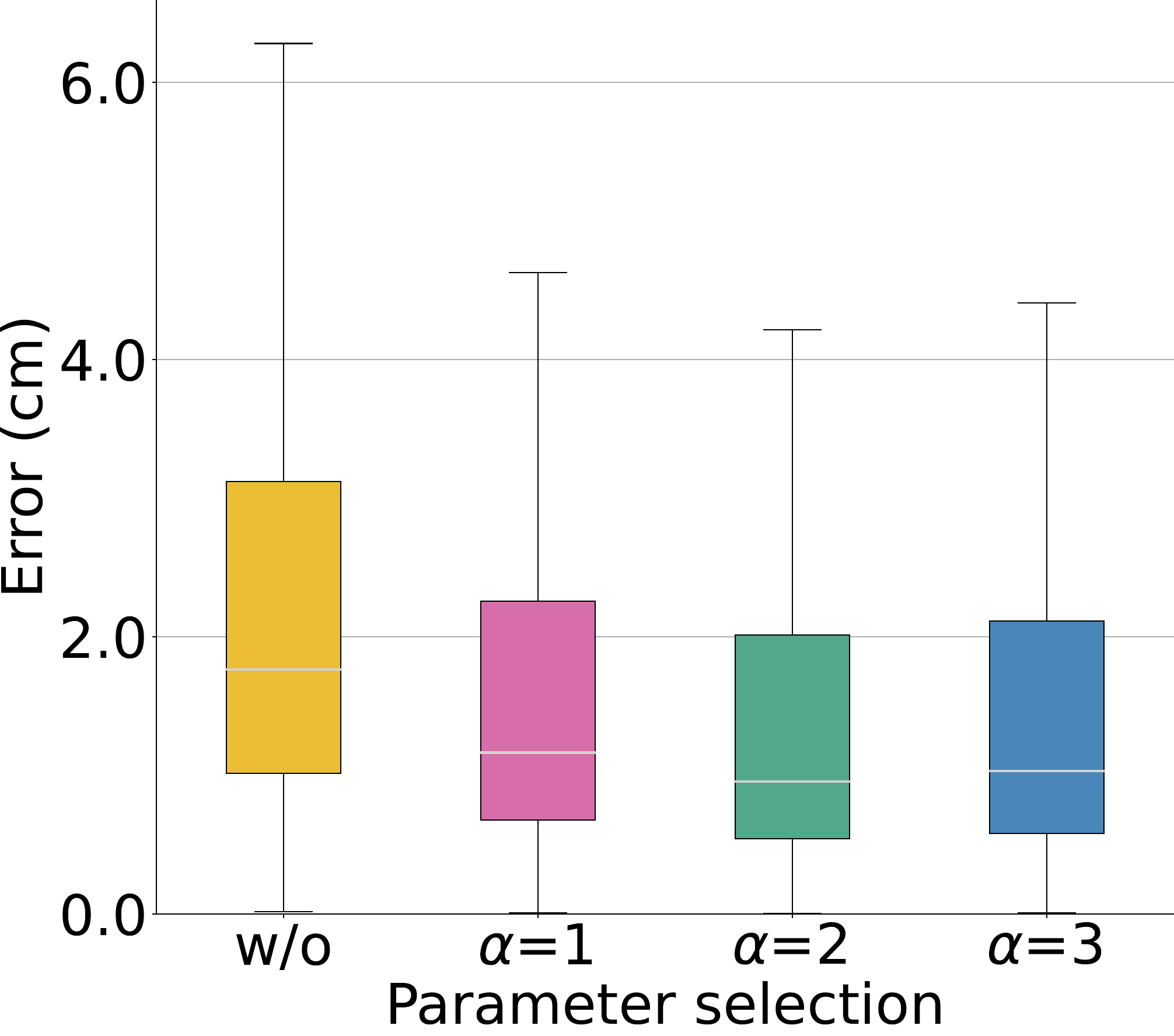}
         \caption{}
         \label{fig:figAblaFCAError}
     \end{subfigure}
        \caption{Quantitative Evaluation of ablation on the frequency controllable attention layer: (a) RMSE of the predicted deformation, (b) error distribution of predicted deformation.}
        \label{fig:figAblaFCAQuant}
\end{figure}

\textit{Frequency-controllable attention layers.} We investigate the impact of FCA layers on the deformation results of both loose and tight-fitting dresses. We approximate the garments' deformations using both the original graph attention network (GAT) without FCA and with FCA using different frequency control parameter values ($\alpha=1$, $2$, and $3$). The qualitative results in Fig.  \ref{fig:figAblaFCA} demonstrate that using GAT without frequency control leads to a loss of high-frequency details in both garments. In contrast, applying FCA layers with $\alpha=2$ and $3$ effectively generates more natural effects. Additionally, we found that the garment deformation around the belly position of sample 1 is better represented in the result for $\alpha=2$. Furthermore, we conduct a quantitative evaluation of the Root Mean Square Error (RMSE) on test data with different FCA settings (Fig. \ref{fig:figAblaFCALoss}). The absence of FCA leads to unstable performance and largest errors, whereas using FCA with frequency control parameter values of $\alpha=2$ and $3$ exhibits similar performance. According to Fig. \ref{fig:figAblaFCAError}, $\alpha=2$ yields slightly better with lower prediction error for our data. Overall, the results provide evidence of the benefits of using the proposed frequency control strategy and selecting suitable parameters to enhance the model's performance in generating realistic garment details. 

\begin{figure*}[t!]
  \centering
  \includegraphics[width= 0.99\linewidth]{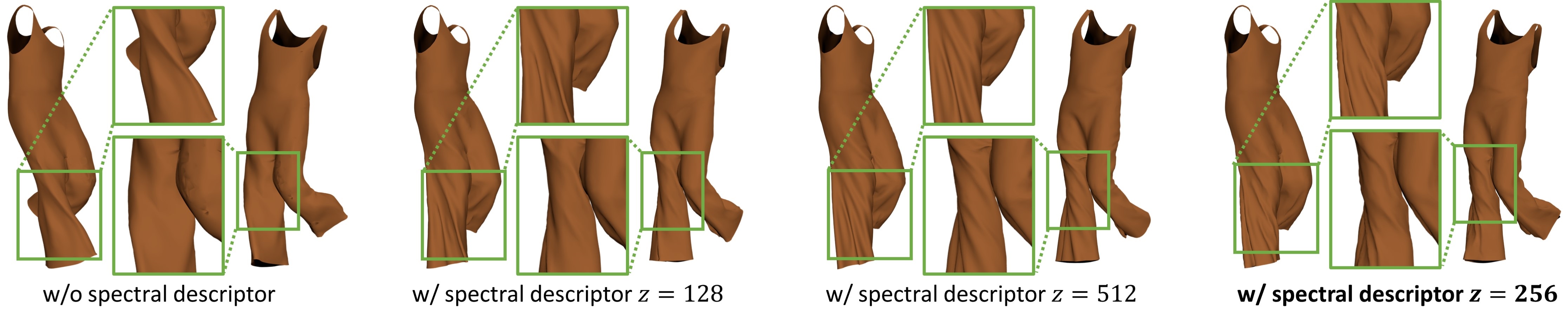}
  \caption{\label{fig:figAblaSpecFeaQual} Qualitative evaluation of ablation on the spectral descriptor. The  trouser leg area are zoomed in on. When without the spectral descriptor, the predictions may have rough and smooth effects that both trouser legs are affected. In contrast, our proposed spectral descriptor improves the above situation to varying degrees for different vector lengths taken.
}
\end{figure*}

\begin{figure}[t!]
     \centering
     \begin{subfigure}[c]{0.23\textwidth}
         \centering
         \includegraphics[width=\textwidth]{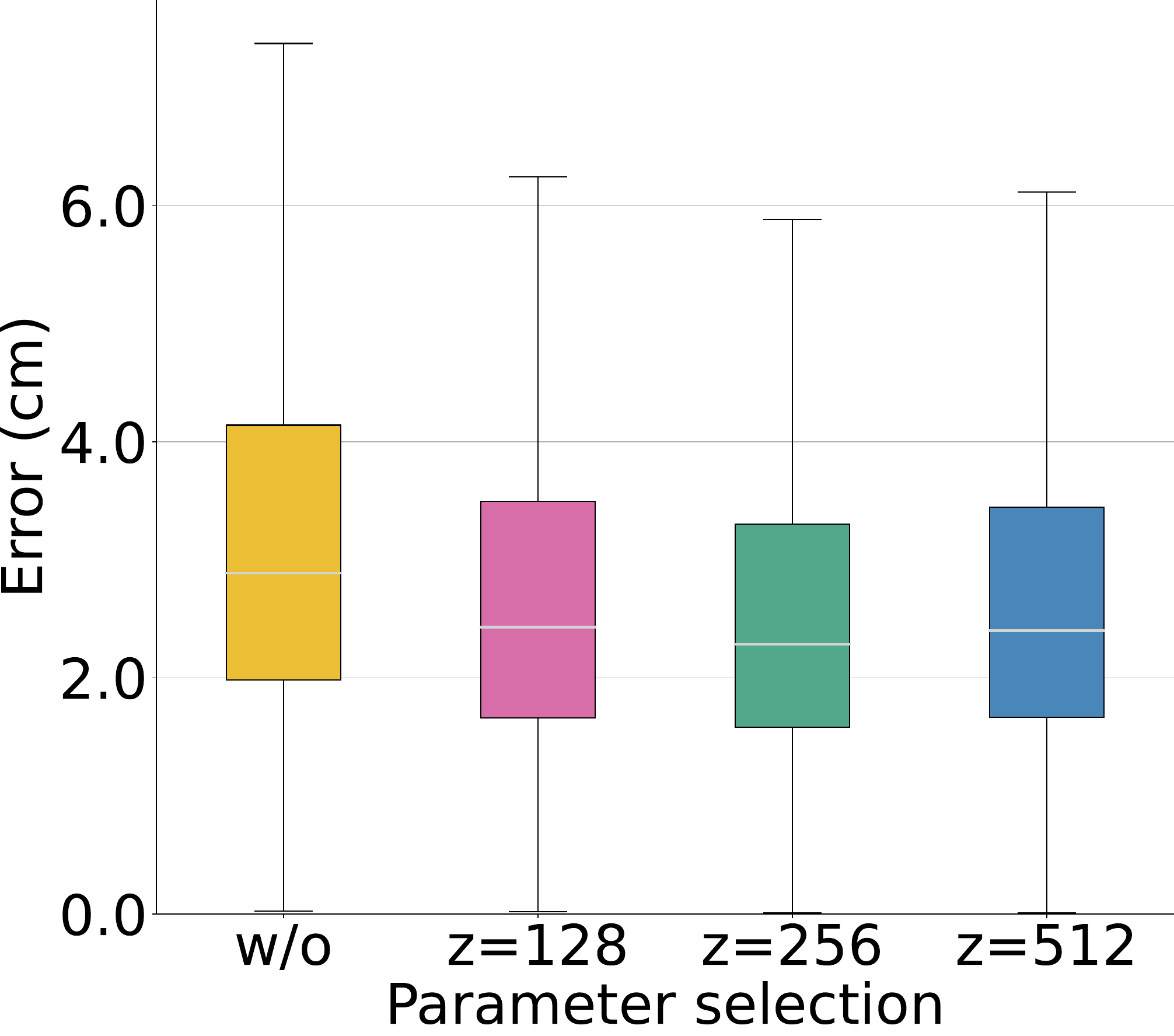}
         \caption{}
         \label{fig:figAblaSpecFeaHard}
     \end{subfigure}
     \hfill
     \begin{subfigure}[c]{0.23\textwidth}
         \centering
         \includegraphics[width=\textwidth]{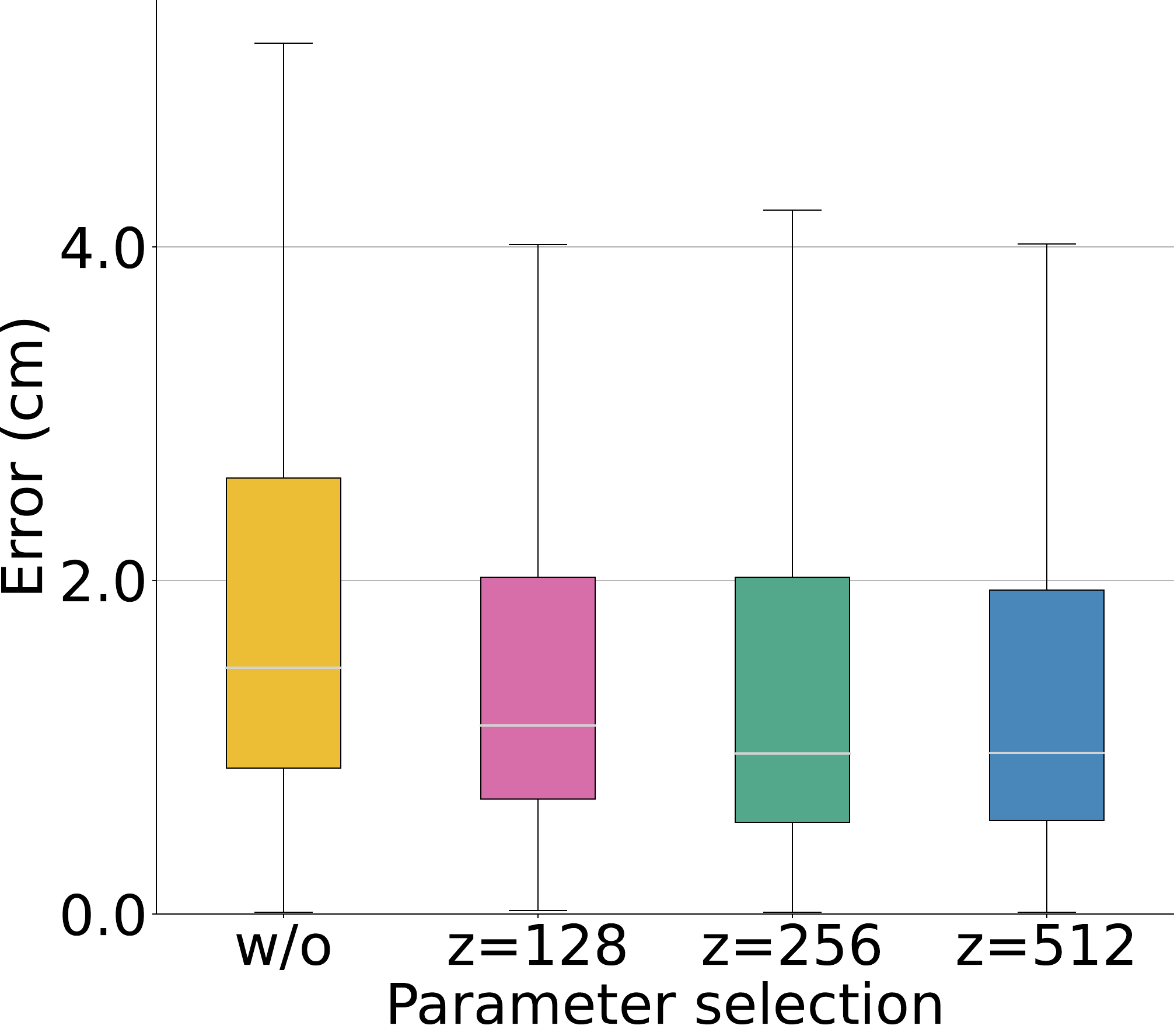}
         \caption{}
         \label{fig:figAblaSpecFea}
     \end{subfigure}
        \caption{Quantitative evaluation of ablation on the spectral descriptor: (a) prediction error of a challenging sample (\textit{i.e.}, jumpsuit), (b) mean error of all test samples. }
        \label{fig:figAblaSpecQuant}
\end{figure}

\textit{Spectral descriptor.} We verify the effectiveness of the proposed spectral descriptor $\boldsymbol{\lambda}=[\lambda_1,...,\lambda_z]$. First, we evaluate its performance against the situation without $\boldsymbol{\lambda}$, specifically using only the position of the body as input to the shape parser. Furthermore, we explore the impact of different descriptor vector lengths where we set $z$ to 128, 256, and 512, and present qualitative results in Fig. \ref{fig:figAblaSpecFeaQual}. Specifically, we test the performance of the model on a challenging garment, a wide-legged jumpsuit, which has fewer trousers and more dresses in our training data. In the leftmost, we observe an unnatural effect around the knee and rougher folds when no spectral descriptor is used.  Despite similar trousers (but not conjoined ones) being present in the training samples, direct positional descriptions do not provide sufficient global information for the model, leading to a prediction failure in regions with unique characteristics. In contrast, the results using $z=256, 512$ demonstrate stronger shape representation capabilities, leading to visually pleasing outputs with finer details. Fig. \ref{fig:figAblaSpecFeaHard} displays the quantitative results, confirming that the proposed spectral descriptor enhances the performance of the model. Specifically, the use of the spectral descriptor significantly reduces the deformation error, with $z=256$ resulting in the lowest error. Fig. \ref{fig:figAblaSpecFea} presents the mean error of all test data, which is approximately 1cm lower than the error of the challenging jumpsuit overall. The results provide support for the effectiveness of the spectral descriptor in accurately deforming dynamic garments.

\begin{figure}[t!]
  \centering
  \includegraphics[width= 1\linewidth]{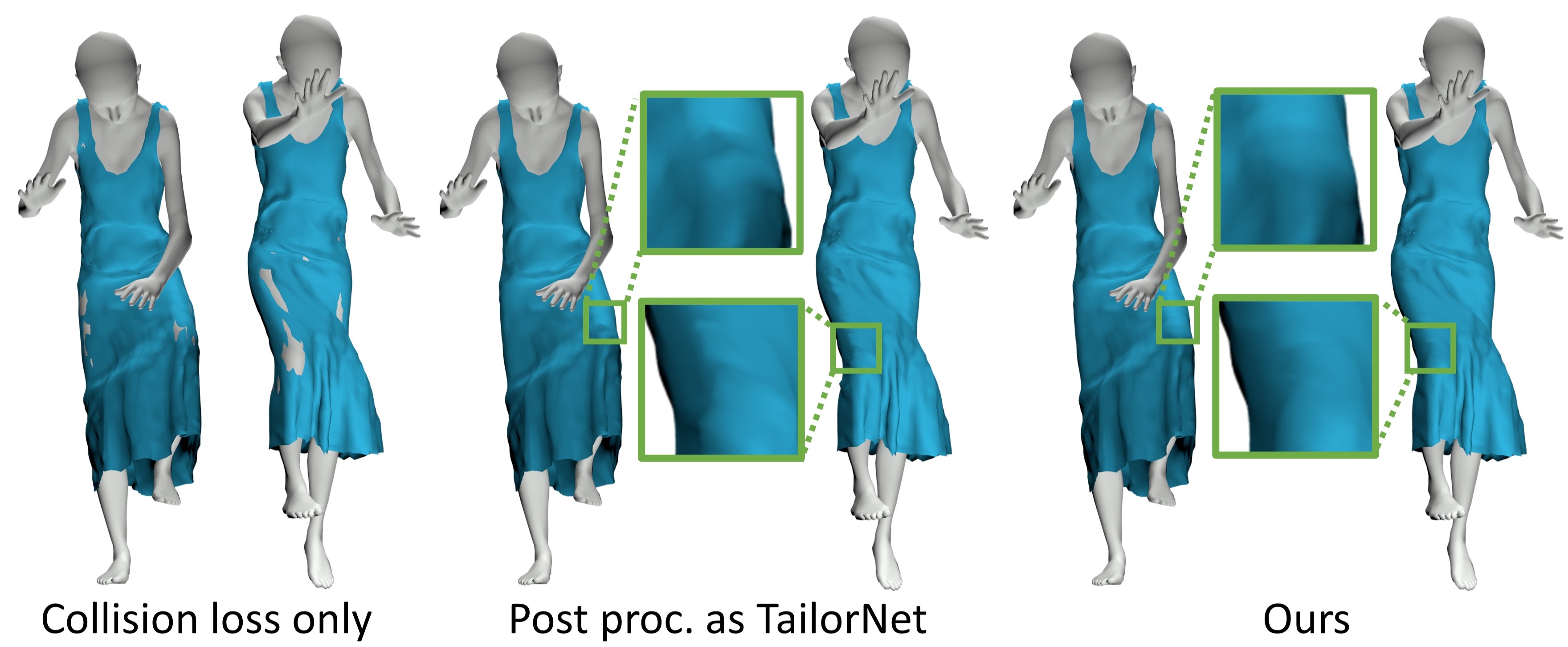}
  \caption{\label{fig:figAblaCollision} Qualitative evaluation of collision handling. We compare our approach with the most common-used collision loss strategies, as well as the post-processing in TailorNet. }
\end{figure}

\textit{Collision handling.} We report the effect of our proposed collision handling solution in Fig \ref{fig:figAblaCollision}. As depicted, while penalizing the penetration of clothing and body imposes a certain degree of constraint on model learning, it cannot ensure a collision-free state in some extreme test poses.  Based on the collision scenario, we implement a post-processing step in TailorNet \cite{TailorNet20} to efficiently resolve collisions. However, this step results in undesirable bulge artifacts, as indicated by the enlarged area. The reason behind these artifacts is that the method crudely employs a fixed distance larger than the penetration depth to remove the vertices of garments that penetrate the body's interior without considering the positions of surrounding vertices. At the rightmost, we present the result produced by our proposed method. By incorporating adaptive adjustments for the collided vertices, the corrected mesh surface appears more natural and achieves higher visual quality. This result validates the effectiveness of our collision handling method, indicating its practical applicability in clothing animation scenarios.

\subsection{Comparisons}
\begin{figure*}[t!]
  \centering
  \includegraphics[width= 0.95\linewidth]{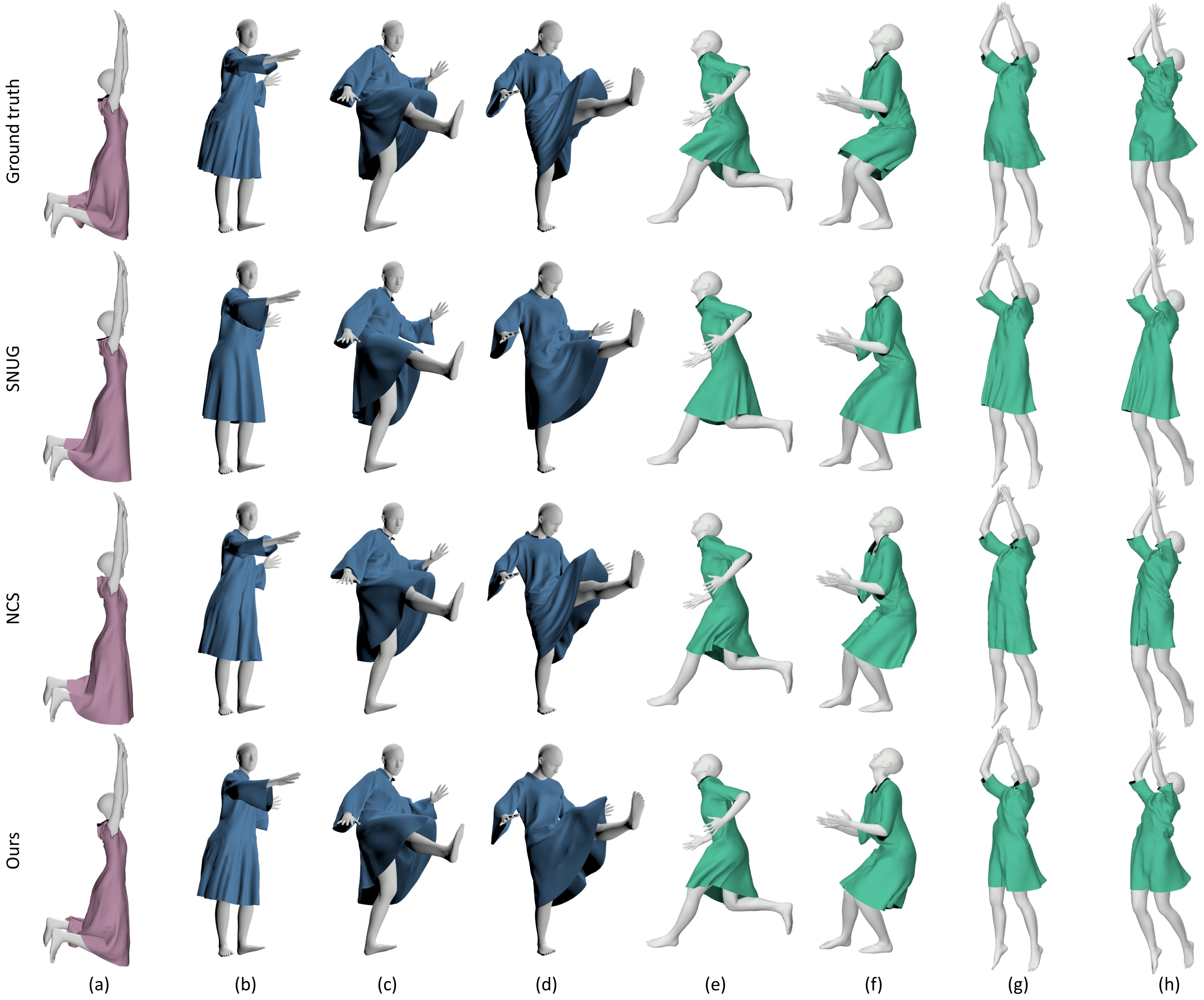}
  \caption{\label{fig:figCompare} Comparison with state-of-art neural dynamic clothing deformation methods. Rows from top to bottom : ground truth, SNUG \cite{SNUG22}, Neural Cloth Simulation (NCS) \cite{Hugo22}, and our prediction. Columns represent various motions: (a) hanging leg raise, (b) palm strike, (c)-(d) kicking with right and left foot, (e)-(h) layup. 
}
\end{figure*}

In Fig. \ref{fig:figCompare}, we compare our method with two recent approaches for dynamic garment deformation, SNUG \cite{SNUG22} and Neural Cloth Simulation (NCS) \cite{Hugo22}. The comparisons include detailed qualitative results for garments in both quasi-static poses (a) and dynamic motions (b)-(h).  Unlike our supervised learning, both SUNG and NCS adopt unsupervised schemes for cloth dynamics by recasting motion equations in physics-based simulation as an optimization problem and applying physics-based loss terms for network training. 
As observed, both SNUG and NCS exhibit visually plausible clothing folds in quasi-static poses, such as the hanging leg raise pose (a). While our method produces reasonably accurate deformations under this pose, there are some subtle differences in details when compared to the ground truth. 
In the case of (b), the character not only moves forward but also twists the body, causing the dress hem to swing back in a continuous motion. However, SNUG does not exhibit such dynamic behavior, particularly in the fold direction of the dress. This lack of dynamics can be attributed to the fact that SNUG is non-dynamic in nature, as fully analyzed in \cite{Hugo22}.  The static behavior also persists in test samples (c)-(h), where the dress does not move naturally with the movement. 
On the other hand, NCS can predict clothing dynamics successfully, where the generated deformations appear consistent with the motions. However, since the method does not require ground truth, the resulting deformations may exhibit unexpected (\textit{i.e.}, far from ground truth) mesh shapes (shown in (h)).  
In contrast, our method demonstrates the accuracy and consistency in generating dynamic clothing deformations, with only a few instances of failure, such as a dynamic failure in predicting the sleeve part in (b) and a detail failure in predicting the wrinkles in (d). Moreover, our method has a key advantage over SNUG and NCS in terms of higher generalization, as it can predict deformations for various garments without repeated training. This is made possible by the proposed spectral strategies and deformation estimator, which enable a unified framework for predicting dynamic garments. 

\section{Conclusion and Future Work}
We have presented a novel approach to efficiently estimate dynamic garment deformation using a unified model. Our work can learn the dynamic behavior of arbitrary garments under consistent body motion, without special constraints on garment topology, vertex count, and degree of fitness. We achieve this by introducing the frequency-control strategies for the deformation network and a spectral global descriptor for diverse garment representation. These techniques enable our deformation estimator to generate personalized and vivid details. We believe that this spectral technique also has the potential to address issues that have been encountered in other animation areas when utilizing graph neural networks. In addition, we propose a neural collision handling method that automatically detects and corrects penetrations between garment-body pairs, resulting in more realistic and natural-looking results. 

There are still a few weaknesses that need to be addressed for further improvement. First, while we were able to mitigate the over-smoothing effect by applying a frequency control strategy to graph attention layers, we did not extend this strategy to other layers in the network, which may have resulted in the spectral bias problem not being fully addressed. In the future, exploring spectral control for the entire network is a promising direction. Second, our current collision handling method relies on the estimated SDF of the human body for detecting and correcting garment penetrations. While we have validated its effectiveness in handling SMPL bodies that were seen during training, the network may not be able to accurately estimate subtle collisions of unseen bodies. Nevertheless, this post-processing step is still necessary and can further improve garment quality compared to only using collision penalties in the learning via loss term. Finally, our method does not account for self-collision cases. Future work could explore ideas for dealing with self-collisions \cite{Tan21}.

{\small
\bibliographystyle{splncs04.bst}
\bibliography{egbib.bib}
}
\end{document}